\documentclass[11pt]{article}

\usepackage[final]{acl}

\usepackage{times}
\usepackage{latexsym}

\usepackage[T1]{fontenc}
\usepackage[utf8]{inputenc}

\usepackage{microtype}

\usepackage{inconsolata}

\usepackage{graphicx}

\usepackage{amssymb}
\usepackage{amsmath}
\usepackage{csquotes}
\usepackage{multirow}
\usepackage{comment}
\usepackage{booktabs}
\usepackage{graphicx}

\title{ReaGeo: Reasoning-Enhanced End-to-End Geocoding with LLMs}

\author{
  \textbf{Jian Cui\textsuperscript{1,2\textdagger}} \quad
  \textbf{Zhiyuan Ren\textsuperscript{1\textdagger}} \quad
  \textbf{Desheng Weng\textsuperscript{1}} \quad
  \textbf{Yongqi Zhao\textsuperscript{1}}
  \\
  \textbf{GONG WENBIN\textsuperscript{1}} \quad
  \textbf{Yu Lei\textsuperscript{1}} \quad
  \textbf{Zhenning Dong\textsuperscript{1,2}}
  \\
  \textsuperscript{1}Amap, Alibaba Group \quad
  \textsuperscript{2}Tsinghua University
  \\
  \texttt{\textsuperscript{1}\{cj87619, renzhiyuan.rzy, wengdesheng.wds, yongqi.zyq\}@alibaba-inc.com}
  \\
  \texttt{\textsuperscript{1}\{gongwenbin.gwb, yu.lei, zhenning.dong\}@alibaba-inc.com}
}

\begin{document}
\maketitle

\renewcommand{\thefootnote}{}
\footnotetext{\textdagger Equal contribution.}

\begin{abstract}
This paper proposes ReaGeo, an end-to-end geocoding framework based on large language models, designed to overcome the limitations of traditional multi-stage approaches that rely on text or vector similarity retrieval over geographic databases, including workflow complexity, error propagation, and heavy dependence on structured geographic knowledge bases. The method converts geographic coordinates into geohash sequences, reformulating the coordinate prediction task as a text generation problem, and introduces a Chain-of-Thought mechanism to enhance the model’s reasoning over spatial relationships. Furthermore, reinforcement learning with a distance-deviation-based reward is applied to optimize the generation accuracy. Comprehensive experiments show that ReaGeo can accurately handle explicit address queries in single-point predictions and effectively resolve vague relative location queries. In addition, the model demonstrates strong predictive capability for non-point geometric regions, highlighting its versatility and generalization ability in geocoding tasks.
\end{abstract}

\section{Introduction}
Geocoding converts human-readable geographic location descriptions into machine-readable coordinates that enable visualization on digital maps, navigation, or spatial positioning \cite{rose2004historical,rushton2007geocoding,goldberg2007text,yin2025toward}. Geocoding demonstrates significant applications across daily life, industrial applications, and academic research. For instance, in traffic accident reporting scenarios, it is necessary to find the accurate geolocations described by the person who made the alarm call \cite{li2020modeling,cheng2024assessing}. Similarly, precise destination identification in logistics order processing benefits from such capabilities. Furthermore, people can identify specific geolocations based on news events or social media messages \cite{zhang2014geocoding,zou2020social}. Industrially, leading platforms such as Google Maps, Microsoft Bing Maps, and Amap (The leading digital map service in China) also use geocoding as a public capability and provide API services\cite{mokhtari2019tagging}. These services support diverse sectors including e-commerce, urban planning, disaster management and so forth, with global API call volumes exceeding millions daily. Consequently, optimizing localization accuracy, such as achieving sub-street-level precision, is essential for geocoding systems.

Traditional geocoding approaches are primarily retrieval-based methods built on structured geographic databases. A common strategy is to decompose an input query into geographic entities using methods such as Named Entity Recognition (NER), hierarchical address tagging, or structural parsing supported by learning-based models \cite{hu2018geo,clemens2020geocoding,radford2021regressing,yin2023chatgpt,li2023enex}. After entity extraction, the parsed components are matched against large-scale geographic databases through rule-based or fuzzy text retrieval mechanisms, often combined with ranking models, spatial constraints, or similarity metrics to link queries with the most plausible geographic entities \cite{melo2017automated,yin2019nlp,yin2019deep,hu2023location,rezaei2024address,yin2025toward}. In addition to text-matching pipelines, an alternative line of work directly embeds address strings into vector representations and performs cosine-similarity retrieval to identify the most likely geographic candidates, which helps improve robustness against spelling variations and informal expressions. Despite their widespread adoption, these pipeline-based methods face several limitations. Incomplete or ambiguous textual descriptions can disrupt entity extraction and degrade retrieval accuracy. Moreover, semantic inconsistencies between user queries and database records—arising from abbreviations, dialectal variations, informal expressions, or inconsistent naming conventions, frequently lead to retrieval failures or incorrect matches.

\begin{figure}[!t]
\centering
\includegraphics[width=\linewidth]{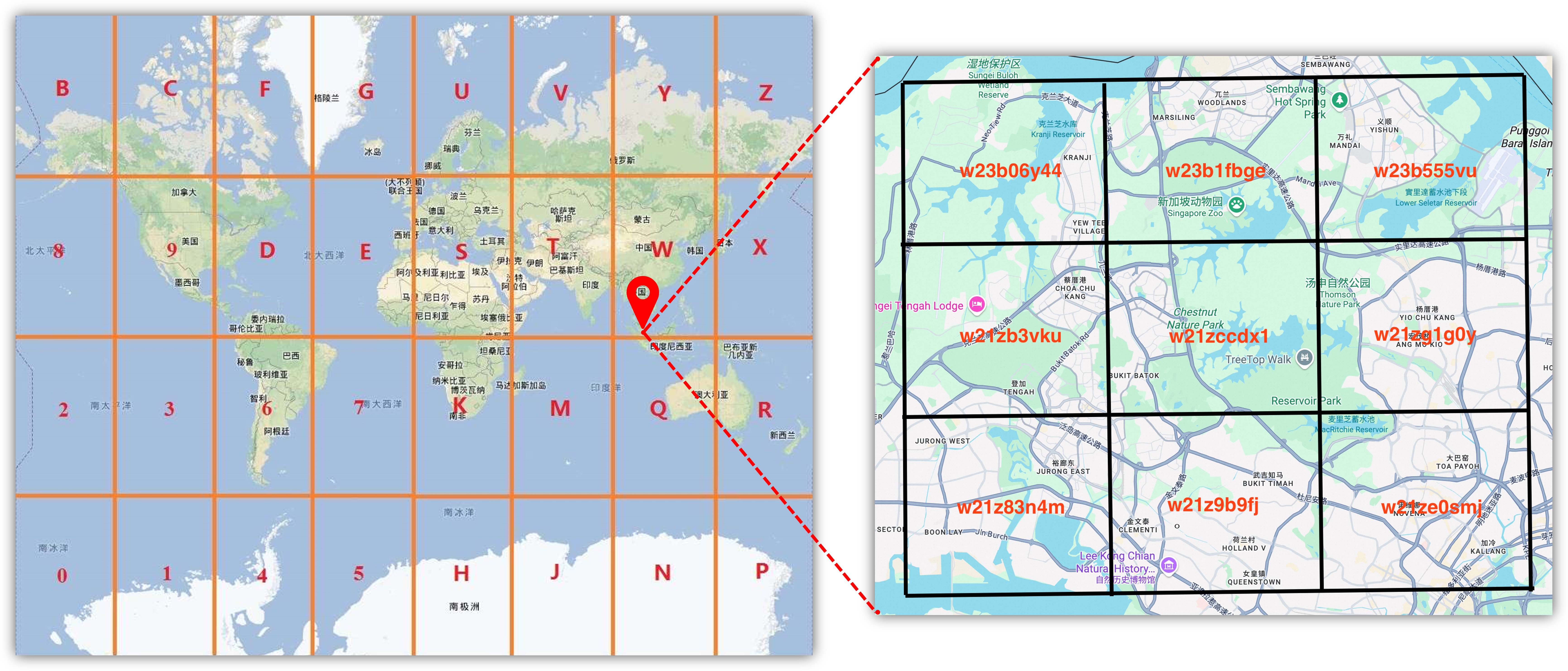}
\caption{Geohash in digital maps. Geohash encodes geographic coordinates into a short alphanumeric string via recursive range bisection and Base32 encoding.}
\label{fig:geohash}
\end{figure}

To address the challenges of traditional pipeline-based geocoding, an innovative end-to-end method based on large language models (LLMs) is adopted, which eliminates the complex and error-prone multi-stage process relying on geographic knowledge bases by directly learning the joint distribution of query texts and geographic spaces. The query text is input to a generative model that auto-regressively decodes a geohash sequence, with the center point of the area represented by the geohash approximated as the final coordinates. Geohash, a spatial indexing method, models the Earth's surface as a two-dimensional plane and recursively partitions it into smaller sub-blocks, with each sub-block assigned the same code within a certain range of latitudes and longitudes \cite{clemens2020geocoding}, as shown in Figure \ref{fig:geohash}. By transforming coordinate prediction into a generative task, alignment between text and geographic modalities is achieved, allowing the powerful language understanding and knowledge memory capabilities of LLMs to be fully leveraged for geocoding applications.

In industrial settings, digital map services such as Google Maps and Amap generate massive volumes of user query and click-through logs daily, which are used to construct an end-to-end training dataset that forms the foundation for large-scale model training. And in this study, Chain-of-Thought (CoT) \cite{wei2022chain} and reinforcement learning are employed to enhance the model’s spatial reasoning and overall prediction performance. Comprehensive experiments demonstrate the superiority of this end-to-end approach, achieving accurate predictions for both single-point locations and non-point geometries.

The main contributions of this paper are summarized as follows:

\begin{itemize}
\item[$\bullet$] ReaGeo, the first end-to-end geocoding model based on LLMs is proposed, offering a unified framework that directly maps textual queries to geohash sequences and thereby eliminates the need for complex and error-prone multi-stage pipelines in geocoding systems.

\item[$\bullet$] A reinforcement learning reward function is developed for geocoding, based on the distance deviation between predicted results and ground truth, which strengthens the model’s spatial reasoning capability and improves its ability to resolve vague or implicitly expressed relative location descriptions.

\item[$\bullet$] The proposed ReaGeo achieves accurate predictions for both single-point locations, including explicit address descriptions and vague relative positional instructions, and non-point geometric shapes, such as line-shaped roads and polygonal administrative areas, demonstrating strong versatility and generalization in complex geospatial tasks.
\end{itemize}

\section{Related Work}
\subsection{Geocoding}
In the field of geocoding, previous research mainly focuses on traditional pipeline-based text retrieval approaches \cite{goldberg2007text,hu2023location,yin2025toward}. These methods typically rely on NER, entity linking, and database queries to perform geocoding \cite{melo2017automated,hu2018geo,mokhtari2019tagging,radford2021regressing,rezaei2024address}. Although these traditional approaches complete geocoding to a certain extent, they face several challenges, such as complex pipelines, dependence on large volumes of manually annotated data, and limited ability to handle ambiguous or incomplete geographic descriptions.

With the rapid development of deep learning and reinforcement learning, the precision and efficiency of geocoding have been significantly improved in recent years. These emerging techniques bring new opportunities to the geocoding field, enabling models to better understand and process complex geographic information. For instance, deep learning automatically extracts salient features from textual and spatial data, while reinforcement learning enhances model performance on complex tasks by optimizing decision-making behaviors \cite{yin2019deep,yin2023chatgpt,li2023enex,rezaei2024address}.

In related research domains, reverse geocoding, which converts coordinates into textual descriptions, also receives increasing attention. Moreover, multi-modal localization studies, such as AddressCLIP \cite{xu2024addressclip}, MixVPR \cite{ali2023mixvpr}, G3 \cite{jia2024g3}, and other works \cite{berton2022rethinking,durgam2024cross,wilson2024image}, explore the integration of visual information for geo-localization. These studies highlight promising directions for future research in leveraging textual, spatial, and visual information to achieve more accurate and robust geocoding.

\subsection{Large Language Models and Reinforcement Learning}
In recent years, LLMs have made significant strides in natural language processing. The introduction of Transformer-based \cite{vaswani2017attention} models, which leverage the self-attention mechanism, dramatically improves machine translation performance and reduces training time. Subsequent architectures such as BERT \cite{devlin2019bert} and GPT \cite{radford2018improving} reshape model pre-training.

GPT-3 \cite{brown2020language}, a model with $175$ billion parameters, achieves notable few-shot learning performance. Techniques like Instruction Fine-Tuning (IFT), Supervised Fine-Tuning (SFT), and Reinforcement Learning from Human Feedback (RLHF) are widely adopted to improve instruction following and alignment with human preferences \cite{ouyang2022training,bai2022training,achiam2023gpt}. Reinforcement learning methods treat an LLM as a policy and optimize task-specific rewards via policy-gradient approaches. Proximal Policy Optimization (PPO) \cite{schulman2017proximal} becomes a de-facto choice by constraining updates within a clipped trust region to keep the language distribution close to a reference model while maximizing reward. Extensions such as Group Relative Policy Optimization (GRPO) \cite{shao2024deepseekmath} compute advantages relative to intra-group baselines to improve sample efficiency and fine-tuning stability for billion-parameter agents. This line of work spawns powerful open-source models (e.g., LLaMA \cite{touvron2023llama}, Qwen \cite{team2024qwen2,yang2024qwen25,yang2025qwen3}, DeepSeek \cite{bi2024deepseek}) that serve as foundations for downstream tasks and for reinforcement learning-based fine-tuning. Reinforcement learning thus plays an increasingly central role in enhancing LLMs’ instruction-following and complex reasoning abilities.

\begin{figure*}[!t]
\centering
\includegraphics[width=\textwidth]{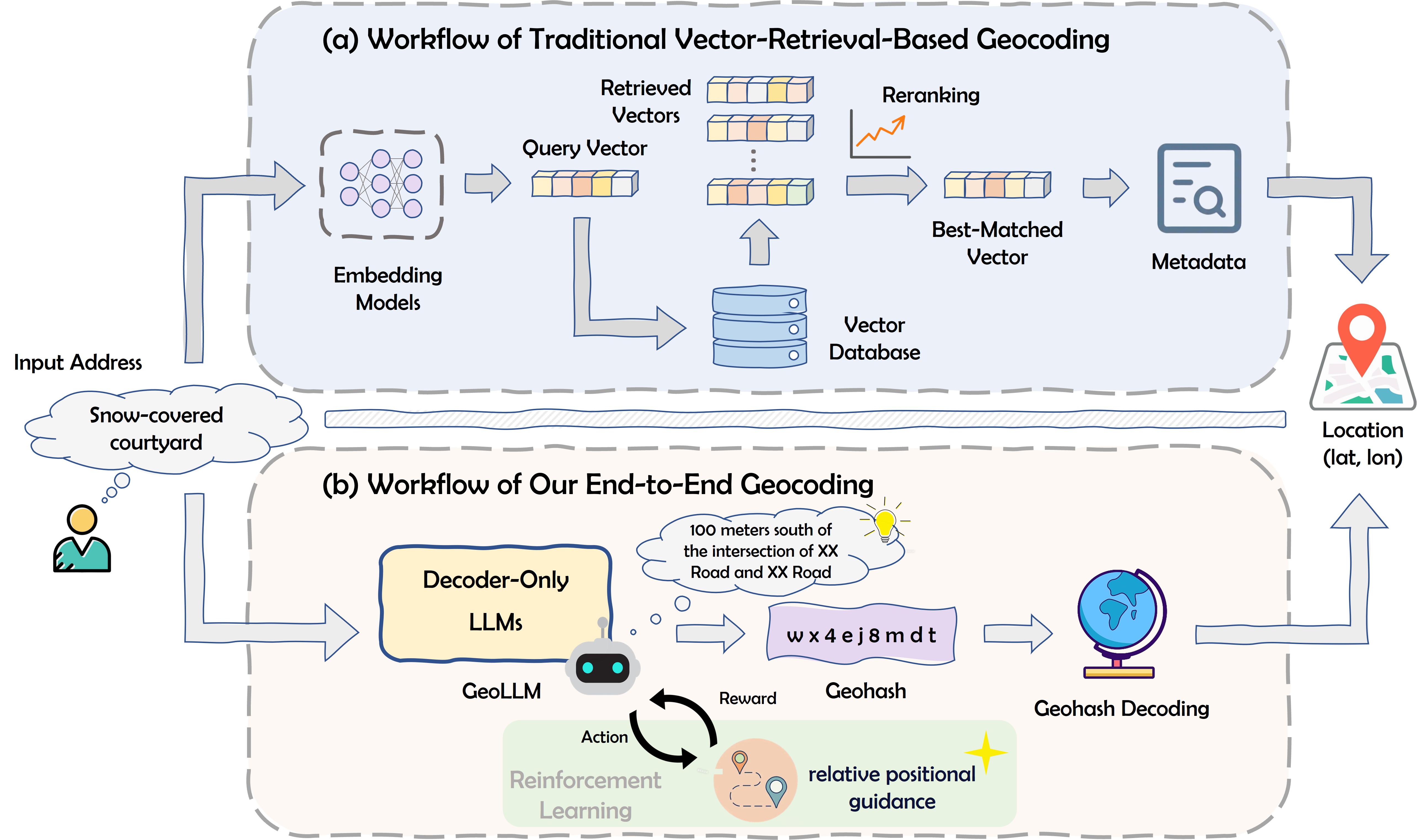}
\caption{The workflow of traditional geocoding methods and our end-to-end geocoding method. (a) Workflow of traditional vector-retrieval-based geocoding: The input address is first transformed into a query vector through an embedding model. This vector is then used to retrieve a set of candidate vectors from a large vector database, followed by a reranking stage to identify the best-matched vector. Finally, associated metadata is retrieved, from which the corresponding geographic coordinates are extracted. (b) Workflow of our end-to-end geocoding: The input query is fed directly into ReaGeo, a decoder-only LLM which is supervised fine-tuned with the CoT method, enabling step-by-step spatial reasoning. To further enhance the model’s capability in resolving relative location descriptions, reinforcement learning with relative positional guidance is employed. ReaGeo generates the geohash sequence in an autoregressive manner, which is subsequently decoded into geographic coordinates.}
\label{fig:architecture}
\end{figure*}

\section{Methods}
\subsection{Preliminaries}
Unlike traditional vector-retrieval-based geocoding methods illustrated in Figure \ref{fig:architecture}(a), the proposed end-to-end approach, ReaGeo, learns the joint distribution between query texts and geographic spaces, thereby inferring the locations that the queries indicate. Formally, this joint distribution constitutes a multi-modal alignment problem, but we convert the geographic space into text through geohash transformation, thus simplifying the multi-modal learning cost. A large language model is adopted as the backbone and is post-trained to complete the alignment of query texts and geohash strings. During inference, the query is fed into ReaGeo, from which the most probable geohash is decoded, and the coordinates of the centroid of the represented region are taken as the final predicted coordinates. This study adopts a geohash Base32 encoding of length $9$, which offers a precision of approximately $2.4$ meters, meeting the accuracy requirements of typical search scenarios. The framework of ReaGeo is illustrated in Figure \ref{fig:architecture}(b), and its geocoding process can be formulated as follows:
\begin{equation}
Y_{\mathrm{geohash}}=F_{\mathrm{ReaGeo}}\left(X_{\mathrm{query}}\right),
\end{equation}
\begin{equation}
y_{\mathrm{pred}}\triangleq
\left(\phi,\lambda\right)=G_{\mathrm{decode}}\left(Y_{\mathrm{geohash}}\right)
\end{equation}
\noindent where $X_{\mathrm{query}}$ denotes the input address query, $F_{\mathrm{ReaGeo}}$ represents the end-to-end mapping performed using GeoLLM, and $Y_{\mathrm{geohash}}$ denotes the output geohash. Subsequently, the geohash decoding $G_{\mathrm{decode}}$ is applied to $Y_{\mathrm{geohash}}$ to obtain the final predicted coordinates $y_{\mathrm{pred}}$, in which $\phi$ is the latitude and $\lambda$ is the longitude, respectively.

The proposed ReaGeo consists of two training stages. In the first stage, the CoT method is introduced to enhance the model’s reasoning capability, thereby capturing the associations among neighboring addresses and improving prediction performance on geographically ambiguous queries. The second stage builds upon the first by employing reinforcement learning with deviation distance as an effective reward signal to further refine positional prediction accuracy.

\subsection{Decoder-Only Architecture}
The core objective of our geocoding task is to directly map natural language queries (such as addresses or location descriptions) into geohash in a generative manner. The decoder-only architecture, by leveraging autoregressive generation, enables end-to-end production of coordinate sequences from textual inputs without relying on complex multi-stage pipeline processes, thereby reducing intermediate error propagation and minimizing dependence on extensive, well-curated geographic knowledge bases.

In this study, Qwen2.5-3B is adopted, which achieves a well-balanced trade-off among language capability, inference accuracy, speed, and resource consumption. This model can effectively handle high-precision parsing of both structured and unstructured address descriptions, while supporting low-latency, high-concurrency deployment in both cloud-based and on-premises production environments.

\subsection{CoT}
To strengthen the association between toponymic references and their formal address representations, thereby improving prediction performance for incomplete or colloquial location descriptions, the CoT method is introduced. Specifically, for a POI address, the approach uses the address itself as the endpoint and its adjacent addresses as the starting points to calculate relative positional relationships. These relative position associations serve as intermediate reasoning steps, extending the model's output to guide it in learning neighborhood knowledge to compensate for the lack of input information before generating the corresponding geohash.

To alleviate memory pressure during inference while maintaining compatibility with the original geohash output format, the CoT framework is combined with Qwen's thinking-mode tagging mechanism. Specifically, special tags \verb|<thinking>| and \verb|</thinking>| are introduced in the training samples to explicitly demarcate intermediate reasoning text, and the thinking mode can be disabled during inference. An example of the data format is as follows:
\begin{verbatim}
"input": "Snow-covered courtyard
<thinking>"
"output": "100 meters south of the
intersection of XX Road and XX Road, XX
District </thinking> w x 4 e j 8 m d t"
\end{verbatim}

\subsection{Reinforcement Learning}
As an efficient and stable policy optimization algorithm, GRPO is widely adopted in various generative tasks. Unlike value-function-based methods such as PPO, GRPO does not require training a separate value model. Instead, it calculates the group relative advantage by normalizing the rewards of a set of candidate outputs generated for the same input query, and then updates the policy based on this relative comparison, thereby significantly reducing training complexity and instability. Based on the above strengths, this study employs GRPO to directly optimize geocoding precision. This approach enables the model to learn fine-grained spatial relationships, especially for understanding precise directional offsets within a query. In this phase, the training data incorporates random offset descriptions, while maintaining the direct query-geohash format. An example of the data format is as follows:
\begin{verbatim}
"input": "50 meters north of No.385 XX
Road, XX Town, XX District, XX"
"output": "w x 4 s f d v p y"
\end{verbatim}

\textit{\textbf{Reward function design.}} For the geocoding task, a task-specific reward function based on geospatial distance deviation is innovatively designed as follows:
\begin{equation}
{R}\left( {{y_{{\rm{pred}}}},{y_{{\rm{true}}}}} \right) = \frac{{\sqrt T  - \sqrt {{D}\left( {{y_{{\rm{pred}}}},{y_{{\rm{true}}}}} \right)} }}{S}
\end{equation}
\noindent where $y_{\mathrm{pred}}$ and $y_{\mathrm{true}}$ are the predicted and ground-truth coordinates, $T$ is a hyperparameter set to $100$, $D\left(\cdot\right)$ computes the WGS-84 ellipsoid-based geodesic distance between two coordinates in meters, and $S$ is a normalization factor set to 1000.

This reward increases as the predicted coordinates approach the true location, encouraging precise spatial inference during training.

\section{Experiments}
In this section, the dataset and evaluation metrics are first introduced, followed by an evaluation of single-point prediction performance on both clear and ambiguous queries, ablation studies to assess the contribution of key components, and finally case studies visualizing predictions for point and non-point geographic regions.

\subsection{Datasets}
A total of $239,918$ data samples from both rural and urban areas of Beijing is obtained via Amap search engine logs and the Amap points of interest (POI) database, and partitioned them into training and test sets. All samples are unique, and for each location in the test set, the corresponding neighboring area is covered within the training set. From the perspective of data distribution and scenario characteristics, there are significant differences between rural and urban areas. Rural areas generally have looser address structures, higher address ambiguity, and sparser infrastructure annotations. In contrast, urban areas have more regular address systems, a high density of POI, and a high degree of road grid formation. These differences directly affect the modeling difficulty and performance of various methods.

For each data sample, two types of input address descriptions are constructed, which are further categorized into two classes of data, as follows:

\textit{\textbf{Base Data.}} Retrieval queries are constructed from three textual sources: search engine queries, POI names, and POI addresses. For search engine queries, the coordinates of the user's final clicked POI are used as the prediction target. For POI names and addresses, the corresponding POI coordinates serve as the prediction target. The training set of the base data is primarily used in ReaGeo’s SFT phase.

\textit{\textbf{Anchor Offset Data.}} Relative location descriptions are commonly used in real-world scenarios such as traffic accident reports and daily communication, making accurate interpretation and geocoding of them highly valuable. To this end, based on the base data, the anchor offset data is constructed, in which phrases such as \enquote{[distance] + meters + [direction] + of} are added to the beginning of each description. The direction is randomly selected from the four cardinal directions (east, south, west, and north), and the distance is uniformly sampled from the range of $30$ to $500$ meters. For example, \enquote{200 meters south of the intersection of XX Road and XX Road}. Meanwhile, the corresponding coordinates are shifted by the distance indicated in its description to generate new labels. The training set of the anchor offset data is primarily used in ReaGeo’s reinforcement learning phase.

The detailed dataset split is presented in Table \ref{tab:dataset}.

\begin{table}[!t]
\caption{Details of Experimental Data Partitioning}
\label{tab:dataset}
\centering
\begin{tabular}{c@{\hspace{10pt}}c@{\hspace{10pt}}c@{\hspace{10pt}}c}
\toprule
Categories & Data & Training & Test \\
\midrule
\multirow{2}{*}{Rural} & Base & 109050 & 11995 \\
& Anchor Offset & 109050 & 11995 \\
\midrule
\multirow{2}{*}{Urban} & Base & 106878 & 11995 \\
& Anchor Offset & 106878 & 11995 \\
\midrule
\multirow{2}{*}{Total} & Base & 215928 & 23990 \\
& Anchor Offset & 215928 & 23990 \\
\bottomrule 
\end{tabular}
\end{table}

\subsection{Implementation Details}
\begin{table*}[!t]
\caption{Quantitative Results of Each Method}
\label{tab:experiment}
\centering
\begin{tabular}{clc@{\hspace{9pt}}c@{\hspace{4pt}}c@{\hspace{4pt}}cc@{\hspace{9pt}}c@{\hspace{4pt}}c@{\hspace{4pt}}c}
\toprule
\multirow{4}{*}{\rotatebox{90}{Data}} & \multirow{4}{*}{Methods} & \multicolumn{4}{c}{Rural} & \multicolumn{4}{c}{Urban} \\
\cline{3-10}
\noalign{\vspace{2pt}}
&& \multirow{2}{*}{ADD} & Acc & Acc & Acc & \multirow{2}{*}{ADD} & Acc & Acc & Acc \\
&&& @100 & @200 & @500 && @100 & @200 & @500 \\
&& (m) & (\%) & (\%) & (\%) & (m) & (\%) & (\%) & (\%) \\
\midrule
\multirow{7}{*}{\rotatebox{90}{Base Data}} & NER+Levenshtein & 5016.7 & 36.2 & 49.2 & 62.4 & 2946.4 & 57.9 & 69.3 & 77.1 \\
& Vector-Retrival@Top1 & 924.1 & 51.6 & 67.7 & 83.4 & 199.1 & 75.3 & 88.7 & 95.6 \\
& Vector-Retrival@Top5+Reranker & 1044.4 & 51.4 & 67.2 & 82.4 & 162.6 & 75.6 & 88.7 & 95.7 \\
& Qwen3-Max & 6548.1 & 0.0 & 1.7 & 8.5 & 1055.0 & 0.0 & 5.1 & 25.4 \\
& Baidu Maps & 2017.5 & 36.0 & 50.3 & 69.4 & 347.7 & 66.4 & 79.8 & 90.1 \\
& Tencent Maps & 1689.2 & 32.8 & 50.9 & 71.9 & 204.1 & 69.9 & 84.9 & 93.7 \\
& ReaGeo (Ours) & \textbf{756.6} & \textbf{58.1} & \textbf{73.5} & \textbf{85.8} & \textbf{119.6} & \textbf{81.9} & \textbf{92.4} & \textbf{97.2} \\
\midrule
\multirow{7}{*}{\rotatebox{90}{Anchor Offset Data}} & NER+Levenshtein & 9422.5 & 4.3 & 12.0 & 38.8 & 6791.2 & 7.3 & 19.4 & 56.3 \\
& Vector-Retrival@Top1 & 1083.5 & 9.8 & 26.5 & 76.9 & 398.4 & 12.7 & 32.6 & 91.2 \\
& Vector-Retrival@Top5+Reranker & 1465.7 & 9.0 & 23.8 & 70.2 & 378.1 & 12.6 & 32.2 & 89.5 \\
& Qwen3-Max & 9207.6 & 0.0 & 0.0 & 5.1 & 1163.1 & 1.7 & 5.1 & 30.5 \\
& Baidu Maps & 2140.4 & 14.6 & 29.7 & 64.6 & 493.5 & 22.1 & 42.2 & 86.2 \\
& Tencent Maps & 1792.0 & 7.3 & 20.3 & 64.7 & 400.0 & 11.8 & 31.2 & 88.5 \\
& ReaGeo (Ours) & \textbf{840.9} & \textbf{23.7} & \textbf{50.6} & \textbf{84.1} & \textbf{208.7} & \textbf{41.5} & \textbf{71.5} & \textbf{96.3} \\
\bottomrule 
\end{tabular}
\end{table*}

In the SFT phase incorporating CoT, a full-parameter fine-tuning of Qwen2.5-3B is conducted on $2$ NVIDIA H20 GPUs, in a distributed data-parallel setup. The batch size is set to $128$ per GPU and the number of epochs to $5$. The AdamW optimizer is employed with an initial learning rate of $5 \times {10^{ - 5}}$.

Reinforcement learning builds upon the SFT-trained checkpoint enhanced with CoT. During GRPO phase, the actor model is updated with a batch size of $512$ prompts per optimization step, subdivided into micro-batches of $128$ to fit memory constraints. Rollout generation is performed with a batch size of $512$ prompts, split into micro-batches of $128$, and each prompt samples $8$ candidate responses to estimate advantages within a group. The number of training epochs in this phase is set to $3$, and the actor learning rate is initialized at $1 \times {10^{ - 6}}$.

\subsection{Evaluation Protocols}
Two metrics are used to evaluate the model's performance, as detailed below:

The first is the average deviation distance (ADD), which measures spatial accuracy by calculating the WGS-84 ellipsoid-based geodesic distance between the predicted coordinates and the ground-truth coordinates, and then averaging the results across all samples. A lower ADD value indicates higher geocoding precision.

The second is a distance-threshold-based accuracy metric (accuracy under distance thresholds, Acc@k), which evaluates the model's localization capability, where $k$ denotes the maximum allowable spatial deviation. Four thresholds are selected, namely $k=100$, $200$, and $500$ meters, represented as Acc@100, Acc@200, and Acc@500, respectively. These metrics indicate the proportion of samples whose predicted coordinates fall within the corresponding distance from the ground-truth coordinates, thereby assessing the model's localization accuracy under varying spatial precision requirements.

\subsection{Quantitative Evaluation}
Quantitative experiments are conducted on rural and urban areas separately using both the base data and the anchor offset data, and the core task is to predict single-point coordinates.

Seven comparison methods are selected, comprising traditional and end-to-end pipelines: 1) NER based on Bert-Base \cite{devlin2019bert} combined with Edit Distance, denoted as \textbf{NER+Levenshtein}; 2) Recalling the single most relevant address using vector-based retrieval with BGE-Base \cite{xiao2024c}, denoted as \textbf{Vector-Retrieval@Top1}; 3) Recalling the top five most relevant addresses using vector-based retrieval with BGE-Base, followed by reranking with BGE-Reranker-Base \cite{xiao2024c}, denoted as \textbf{Vector-Retrieval@Top5+Reranker}; 4) \textbf{Qwen3-Max} \cite{yang2025qwen3}, the Qwen series model with the best overall performance to date, without any fine-tuning; 5) the geocoding API of Baidu Maps \cite{baidu2025geocoding}, denoted as \textbf{Baidu Maps}; and 6) the geocoding API of Tencent Maps \cite{tencent2025geocoding}, denoted as \textbf{Tencent Maps}.

\subsubsection{Performance on Base Data}
The experimental results are shown in Table \ref{tab:experiment}. In both rural and urban areas, NER+Levenshtein exhibits significant limitations, primarily due to the complex structure of geographic addresses and it's inability to capture semantic equivalence. Vector-retrieval-based methods are relatively stable and achieve decent performance, yet they are still constrained by the coverage of the retrieval database and the accuracy of ranking. Qwen3-Max predicts geographic coordinates based on its existing knowledge, but even though the search range is constrained, the results remain highly unsatisfactory. Baidu Maps and Tencent Maps are limited by their matching mechanisms and data coverage, resulting in only mediocre performance when handling ambiguous or non-standard addresses.

ReaGeo outperforms others, achieving the lowest ADD and the highest Acc@k on both rural and urban data. It's generalized understanding allows it to use numerous address annotations for accurate alignment and reason about semantic equivalence through pre-trained geographical knowledge. Furthermore, It leverages LLMs' long-tail information understanding to handle vague addresses and strong context dependence, and its joint modeling of geographical prior and dynamic reasoning enables reasoning about implicit spatial relationships.

\subsubsection{Performance on Anchor Offset Data}
As presented in Table \ref{tab:experiment}, all the methods exhibit varying degrees of performance degradation on the anchor offset data. The ADD metric degrades particularly severely for NER+Levenshtein and Qwen3-Max. For the former, the randomly added positional bias descriptions in the addresses increase the edit distance, making it more likely to miss the optimal match. For the latter, such random positional bias descriptions introduce greater confusion into the model’s reasoning process. As for the other baseline methods, their performance degradation falls within expected ranges, and their overall performance remains suboptimal.

Across both rural and urban scenarios in the anchor offset data, ReaGeo demonstrates remarkable robustness and precision. It still outperforms other methods due to its understanding of complex geographical semantics and innovative training strategies that target core geocoding challenges like address ambiguity, spatial relationship reasoning, and distribution shift.

To further demonstrate the advantages of the proposed method, experiments are conducted on a second dataset, which will be publicly released, with details provided in Appendix \ref{sec:singapore}.

\begin{table*}[!t]
\caption{Ablation Study of Key Components}
\label{tab:ablation}
\centering
\begin{tabular}{lccccc}
\toprule
Methods & ADD (m) & Acc@100 (\%) & Acc@200 (\%) & Acc@500 (\%) & EC \\
\midrule
w/o CoT & 475.5 & 48.7 & 69.3 & 90.0 & 80 \\
w/o thinking delimiters & \textbf{474.2} & 50.7 & 71.7 & 90.6 & 21 \\
w/o GRPO & 599.2 & 40.6 & 57.0 & 86.1 & 26 \\
full & 481.2 & \textbf{51.3} & \textbf{72.0} & \textbf{90.9} & \textbf{20} \\
\bottomrule 
\end{tabular}
\end{table*}

\begin{table}[!t]
\caption{Comparison of Performance under Different Output Formats in SFT}
\label{tab:ablation_coordinates}
\centering
\begin{tabular}{lcccc}
\toprule
\multirow{3}{*}{Methods} & \multirow{2}{*}{ADD} & Acc & Acc & Acc \\
&& @100 & @200 & @500 \\
& (m) & (\%) & (\%) & (\%) \\
\midrule
Coordinates & 440.1 & 68.7 & 81.8 & 90.6 \\
Geohash & \textbf{427.7} & \textbf{68.8} & \textbf{82.5} & \textbf{91.5} \\
\bottomrule 
\end{tabular}
\end{table}

\begin{table}[!t]
\caption{Comparison of Performance between Cardinal and Unseen Intercardinal Directions}
\label{tab:ablation_directions}
\centering
\begin{tabular}{lcccc}
\toprule
\multirow{3}{*}{Methods} & \multirow{2}{*}{ADD} & Acc & Acc & Acc \\
&& @100 & @200 & @500 \\
& (m) & (\%) & (\%) & (\%) \\
\midrule
Cardinal & 524.8 & 32.6 & 61.1 & 90.2 \\
Intercardinal & 562.9 & 22.4 & 50.5 & 89.4 \\
\bottomrule 
\end{tabular}
\end{table}

\subsection{Ablation Studies}
The effectiveness of the proposed method is validated from four aspects.

\textit{\textbf{Analysis of Key Components.}} Here, the base data and the anchor offset data are combined. Table \ref{tab:ablation} presents the results of the ablation study on key components. Removing the reinforcement learning component leads to a significant degradation in performance metrics, highlighting its critical role in enhancing geocoding precision. Eliminating the CoT mechanism also leads to degraded geocoding performance and an increased number of instances in which the model produces invalid outputs (referred to as error count, EC). Additionally, the thinking delimiters within the CoT mechanism offer a minor improvement in the proportion of high-precision samples.

\begin{figure}[!t]
\centering
\includegraphics[width=\linewidth]{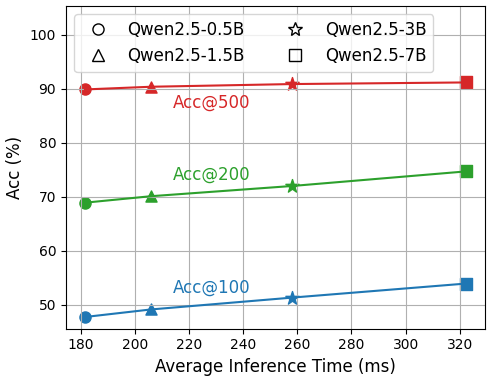}
\caption{Comparison of Qwen with different model sizes. Both inference accuracy and latency on an NVIDIA H20 GPU increase as model size grows.}
\label{fig:ablation}
\end{figure}

\textit{\textbf{Impact of Model Size.}} Experiments are also conducted on the combined base and anchor offset data here. Figure \ref{fig:ablation} compares the geocoding performance of Qwen with different model sizes. As the model size increases, localization accuracy continuously improves, yet the average inference time also grows. The Qwen2.5-3B strikes a good balance between accuracy and efficiency, making it the preferred choice for practical applications.

\textit{\textbf{Comparison of Output Format.}} On the base data, the Qwen2.5-3B model is evaluated under SFT, comparing direct generation of latitude–longitude coordinates with geohash sequences. Table \ref{tab:ablation_coordinates} demonstrates that geohash discretizes continuous geographic space into a sequential format, which better aligns with the text generation mechanism of language models and supports progressive precision control.

\textit{\textbf{Generalization to Intercardinal Directions.}} An anchor-offset test dataset with four intercardinal directions (northeast, southeast, northwest, and southwest) is constructed, and the performance of the proposed method on it is compared with that on the previously used four cardinal directions (east, west, south, and north), as shown in Table \ref{tab:ablation_directions}. It can be observed that the model maintains relatively high accuracy even on unseen intercardinal directions, demonstrating a certain level of generalization capability.

\subsection{Case Studies}
In the case studies, concrete examples are used to demonstrate that ReaGeo, while performing single-location prediction, shows strong inference capability for directional and distance indicators. Additionally, it can estimate the approximate extent of non-point geometries such as lines and polygons.

\subsubsection{Single-Point Prediction}
\begin{figure}[!t]
\centering
\includegraphics[width=\linewidth]{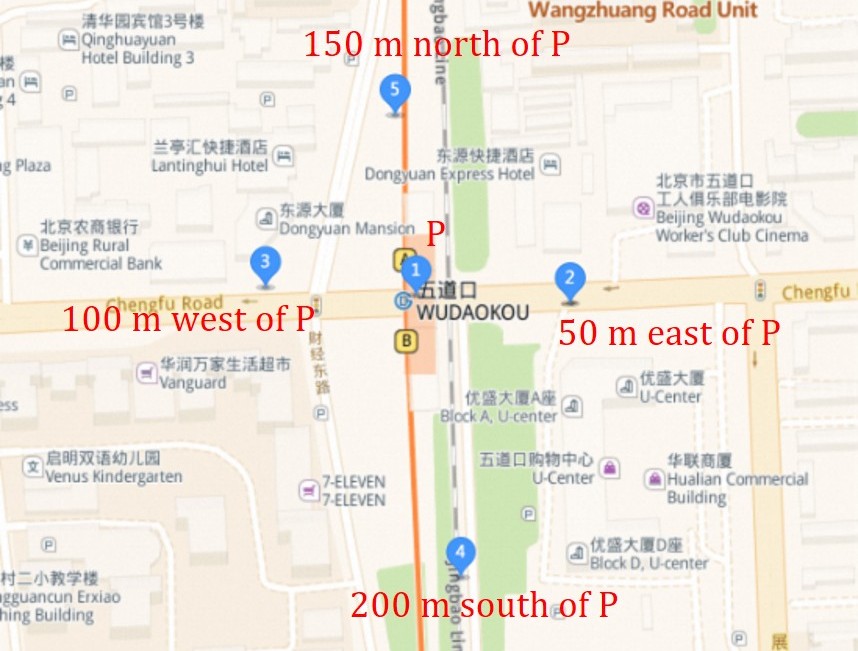}
\caption{Visualization of single-point location prediction. The markers 2–5 correspond to predicted positions guided by relative-position cues around landmark 1.}
\label{fig:single_point}
\end{figure}

To evaluate ReaGeo’s ability to handle relative location descriptions, it's asked to predict the locations of points at specified offsets in the four cardinal directions relative to a given landmark (e.g., Wudaokou Subway Station, denoted as point P). The prediction results, visualized in Figure \ref{fig:single_point}, show that ReaGeo can effectively estimate the relative positional shifts of these points around the landmark within an acceptable error range, demonstrating notable practical value.

\subsubsection{Non-Point Geometry Prediction}
To enable non-point region prediction, beam search with a beam width of $50$ is applied to each POI. Multiple candidate coordinate points are generated through this process. These points are then visualized on a digital map to intuitively show the spatial distribution of the predictions. This strategy offers a more comprehensive depiction of predictive uncertainty and visualizes the spatial patterns of the results clearly. The following renders the predicted line and polygon POIs.

\begin{figure}[!t]
\centering
\includegraphics[width=\linewidth]{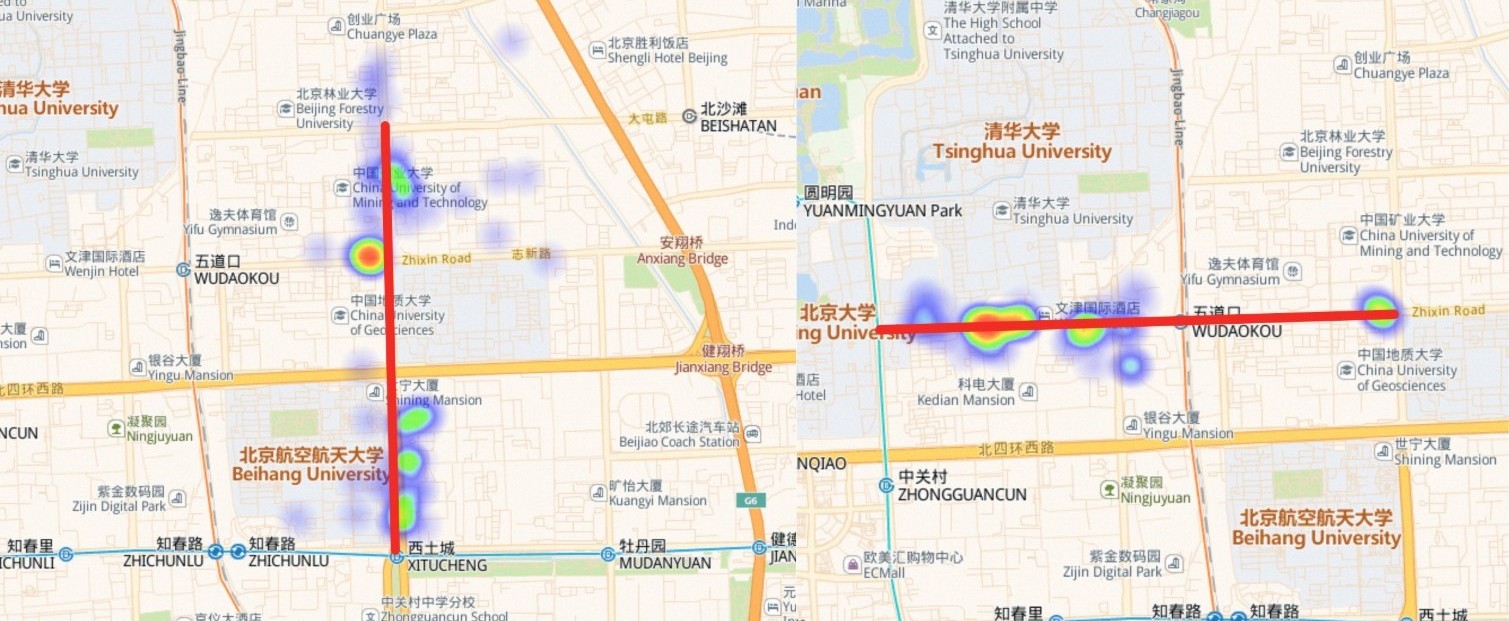}
\caption{Visualization of line POI prediction. Coordinate points are rendered along road segments.}
\label{fig:road}
\end{figure}

\textit{\textbf{Line POI Prediction.}} As illustrated by examples in Figure \ref{fig:road}, the heat map visualization shows rendering points distributed along both roads (red lines), revealing distinctive distribution patterns. Although some rendering points deviate from the road centerlines and outlines are incomplete, they still generally follow the roads, providing an initial view of the heat distribution along the roads.

\begin{figure}[!t]
\centering
\includegraphics[width=\linewidth]{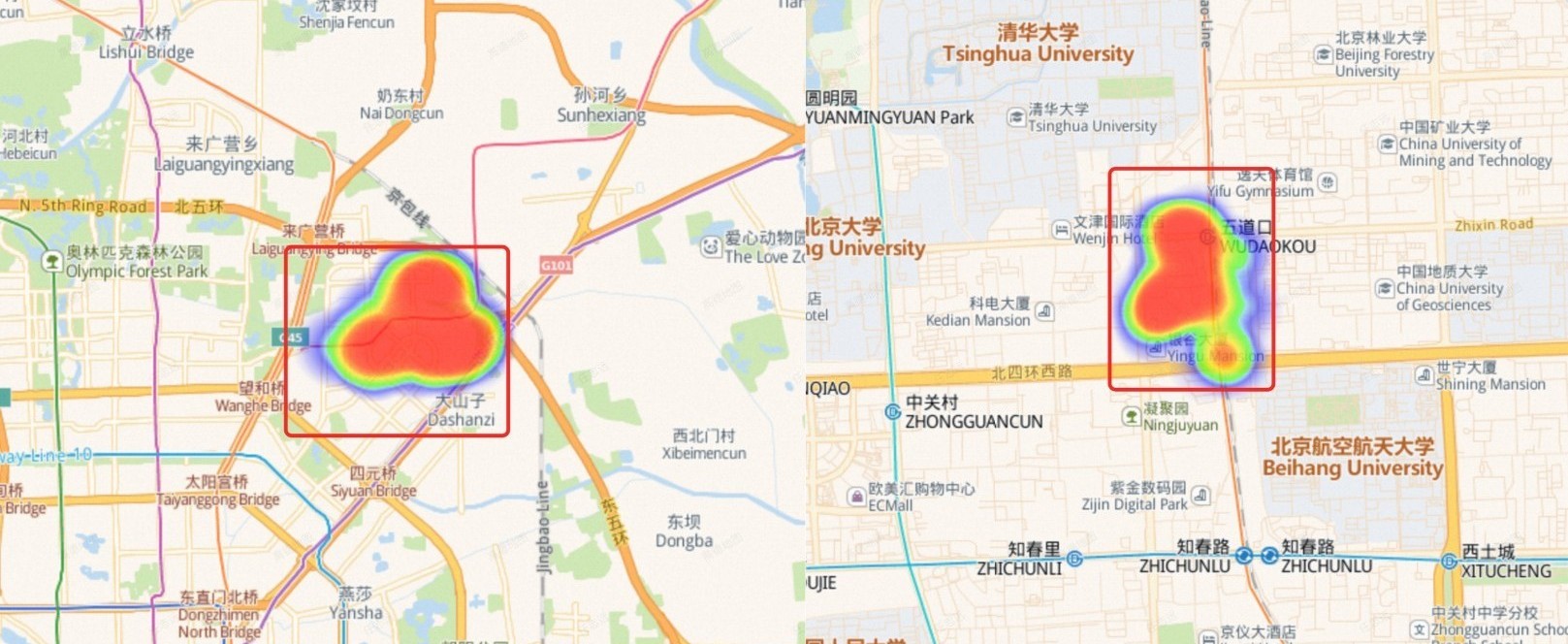}
\caption{Visualization of polygon POI prediction. The heat map shows a polygon area with its administrative boundary, with high-intensity rendering concentrated in the core zone and surrounding regions less highlighted.}
\label{fig:polygon}
\end{figure}

\textit{\textbf{Polygon POI Prediction.}} As illustrated by examples in Figure \ref{fig:polygon}, the heat map rendering shows the distribution of activity within each district. The red outlines mark the administrative boundaries of the respective areas, while the heat map visualizes the actual concentration of points. It can be seen that the heat map coverage accounts for roughly $70\%$ of each district. Upon closer inspection, the heat rendering is mainly concentrated in the core zones, while other parts, although still within the administrative boundaries, are not directly highlighted.

\section{Conclusions}
This study introduces ReaGeo, an end-to-end geocoding framework that utilizes LLMs, CoT and reinforcement learning to directly convert textual location descriptions into geohash. The proposed method simplifies the traditional multi-stage pipeline and demonstrates robust performance in handling both explicit addresses and vague relative location queries. It also demonstrates a promising capability to predict non-point geographic regions, including lines and polygons, indicating significant potential for handling complex geospatial data types. Subsequent work will focus on multi-modal geocoding by integrating visual data.

\section*{Limitations}
Although the proposed end-to-end geocoding framework achieves a higher upper bound of accuracy, it still has notable limitations. The current experiments are limited to a single city, and achieving localization over a larger geographic area requires a larger language model, which comes at the cost of increased computational resource consumption. In future work, the scalability of the model across hundreds of cities will be investigated, and the trade-off between backbone model size and geographic coverage breadth will be examined. Additionally, in the proposed LLM-based geocoding system, disentangling memorization from reasoning remains an open problem that warrants further investigation.

\section*{Acknowledgements}
This research was supported by the National Natural Science Foundation of China (NSFC) under Grants 72571008.

% Bibliography entries for the entire Anthology, followed by custom entries
%\bibliography{custom,anthology-overleaf-1,anthology-overleaf-2}

% Custom bibliography entries only
\bibliography{references}

\begin{table*}[!t]
\caption{Quantitative Results of Each Method on the Singapore Dataset}
\label{tab:singapore_experiment}
\centering
\begin{tabular}{cl@{\hspace{18pt}}cccc}
\toprule
\multirow{2}{*}{Data} & \multirow{2}{*}{Methods} & ADD & Acc@100 & Acc@200 & Acc@500 \\
&& (m) & (\%) & (\%) & (\%) \\
\midrule
\multirow{4}{*}{\shortstack{Base \\ Data}} & NER+Levenshtein & 4347.6 & 24.1 & 33.1 & 46.1 \\
& Vector-Retrival@Top1 & 2964.6 & 54.0 & 60.8 & 68.1 \\
& Vector-Retrival@Top5+Reranker & 2790.5 & 59.2 & 64.4 & 70.7 \\
& ReaGeo (Ours) & \textbf{2152.3} & \textbf{65.9} & \textbf{69.6} & \textbf{74.2} \\
\midrule
\multirow{4}{*}{\shortstack{Anchor \\ Offset \\ Data}} & NER+Levenshtein & 5826.0 & 3.7 & 10.6 & 33.5 \\
& Vector-Retrival@Top1 & 3455.6 & 7.6 & 19.7 & 59.4 \\
& Vector-Retrival@Top5+Reranker & 3662.6 & 7.8 & 19.7 & 58.8 \\
& ReaGeo (Ours) & \textbf{2395.9} & \textbf{10.6} & \textbf{25.3} & \textbf{70.8} \\
\bottomrule 
\end{tabular}
\end{table*}

\newpage
\appendix

\section{Quantitative Evaluation on a Second Dataset}
\label{sec:singapore}

\begin{table}[!t]
\caption{Details of Data Partitioning for the Singapore Dataset}
\label{tab:singapore_dataset}
\centering
\begin{tabular}{cccc}
\toprule
Data & Training & Test & Total \\
\midrule
Base & 60,000 & 6,669 & 66,669 \\
Anchor Offset & 60,000 & 6,669 & 66,669 \\
\bottomrule 
\end{tabular}
\end{table}

To further demonstrate the advantages of the proposed method, an additional dataset covering the city of Singapore is constructed. The dataset will be publicly released on \url{https://github.com/ziyue246/ReaGeo} to facilitate future research. As the second dataset in the experiments, it follows the same partitioning strategy as the main dataset, consisting of the base data and the anchor-offset data. Each subset contains $60,000$ training samples and $6,669$ test samples, as presented in Table \ref{tab:singapore_dataset}.

The quantitative evaluation results of each method on the Singapore dataset are shown in Table \ref{tab:singapore_experiment}. Notably, some methods are not included in the tables because Qwen3-Max largely relied on guesswork, resulting in substantial errors, and the APIs of Baidu Maps and Tencent Maps do not support English addresses. The experimental results indicate that the proposed end-to-end method still outperforms other retrieval-based approaches in terms of performance metrics.

\end{document}